\documentclass[sigconf]{acmart}
\AtBeginDocument{%
  }

\setcopyright{cc}
\copyrightyear{2025}
\acmYear{2025}
\acmDOI{10.1145/3765612.3767226}
\acmConference[BCB '25]{16th ACM International Conference on Bioinformatics, Computational Biology and Health Informatics}{October 11--15, 2025}{Philadelphia, PA}
\acmISBN{978-1-4503-XXXX-X/2025/11}

\usepackage{enumitem}
\setlist[itemize]{nosep, topsep=4pt, partopsep=0pt, parsep=0pt, leftmargin=*}
\setlist[enumerate]{nosep, topsep=4pt, partopsep=0pt, parsep=0pt, leftmargin=*}

\usepackage[table,xcdraw]{xcolor}
\usepackage{multirow}

\begin{document}

\title{ATHENA: Atherosclerosis through Hierarchical Explainable Neural Network Analysis}


\author{Irsyad Adam}
\affiliation{%
  \institution{University of California, Los Angeles}
  \city{Los Angeles}
  \country{United States}}
\email{irsyadadam@g.ucla.edu}

\author{Steven Swee}
\affiliation{%
  \institution{University of California, Los Angeles}
  \city{Los Angeles}
  \country{United States}}
\email{sswee@g.ucla.edu}

\author{Erika Yilin Zheng}
\affiliation{%
  \institution{University of California, Los Angeles}
  \city{Los Angeles}
  \country{United States}}
\email{erikayilin@g.ucla.edu}

\author{Ethan Ji}
\affiliation{%
  \institution{University of California, Los Angeles}
  \city{Los Angeles}
  \country{United States}}
\email{eji@g.ucla.edu}

\author{William Speier}
\affiliation{%
  \institution{University of California, Los Angeles}
  \city{Los Angeles}
  \country{United States}}
\email{speier@g.ucla.edu}

\author{Dean Wang}
\affiliation{%
  \institution{University of California, Los Angeles}
  \city{Los Angeles}
  \country{United States}}
\email{dingwang@g.ucla.edu}

\author{Alex Bui}
\affiliation{%
  \institution{University of California, Los Angeles}
  \city{Los Angeles}
  \country{United States}}
\email{buia@mii.ucla.edu}

\author{Wei Wang}
\affiliation{%
  \institution{University of California, Los Angeles}
  \city{Los Angeles}
  \country{United States}}
\email{weiwang@cs.ucla.edu}

\author{Karol Watson}
\affiliation{%
  \institution{University of California, Los Angeles}
  \city{Los Angeles}
  \country{United States}}
\email{kwatson@mednet.ucla.edu}

\author{Peipei Ping}
\affiliation{%
  \institution{University of California, Los Angeles}
  \city{Los Angeles}
  \country{United States}}
\email{pping38@g.ucla.edu}

\renewcommand{\shortauthors}{Adam et al.}
\begin{abstract}
In this work, we study the problem pertaining to personalized classification of subclinical atherosclerosis by developing a hierarchical graph neural network framework to leverage two characteristic modalities of a patient: clinical features within the context of the cohort, and molecular data unique to individual patients. Current graph-based methods for disease classification detect patient-specific molecular fingerprints, but lack consistency and comprehension regarding cohort-wide features, which are an essential requirement for understanding pathogenic phenotypes across diverse atherosclerotic trajectories. Furthermore, understanding patient subtypes often considers clinical feature similarity in isolation, without integration of shared pathogenic interdependencies among patients. To address these challenges, we introduce ATHENA: \textbf{A}therosclerosis \textbf{T}hrough \textbf{H}ierarchical \textbf{E}xplainable \textbf{N}eural Network \textbf{A}nalysis, which constructs a novel hierarchical network representation through integrated modality learning; subsequently, it optimizes learned patient-specific molecular fingerprints that reflect individual omics data, enforcing consistency with cohort-wide patterns. With a primary clinical dataset of 391 patients (PESA study), their respective transcriptomics signatures, as well as their STRING PPI profiles, we demonstrate that this heterogeneous alignment of clinical features with molecular interaction patterns has significantly boosted subclinical atherosclerosis classification performance across various baselines by up to 13\% in area under the receiver operating curve (AUC) and 20\% in F1 score. We further validated ATHENA on a secondary clinical dataset on atherosclerosis (by Steenman and Espitia et al.). Taken together, ATHENA enables mechanistically-informed patient subtype discovery through explainable AI (XAI)-driven subnetwork clustering; this novel integration framework strengthens personalized intervention strategies, thereby improving the prediction of atherosclerotic disease progression and management of their clinical actionable outcomes.

\end{abstract}




\received{10 September 2025}

\maketitle

\section{Introduction}
Subclinical atherosclerosis is a widespread yet insidious precursor to life-threatening cardiovascular events \cite{Ullah, Martin, Torrijo, Barkas}. The accumulation of plaque in the arteries and resulting carotid intima-media thickness have been applied as indicators of atherosclerotic pathogenesis \cite{Ullah, Martin}. These markers remain imperceptible to patients until occlusion or rupture occurs. Noncontrast CT and other imaging platforms have been widely implemented for the diagnosis and risk stratification of subclinical atherosclerosis \cite{Ross, Oleary, Peters}. However, these approaches remain operator-dependent and thus, inherently inconsistent, limiting their applications to clinical care \cite{Papageorgiou}. Given the silent nature of this disease, early detection of plasma molecular markers is imperative to guiding tailored intervention before irreversible damage ensues. Molecular profiling is valuable in quantifying patient-specific atherosclerotic burden, helping stage or subtype a patient. However, leveraging specific omics data (e.g. transcriptomics) to better understand clinical subtypes of atherosclerosis remains a challenge in the field.  

\begin{figure}
    \includegraphics[width=0.4\textwidth, keepaspectratio]{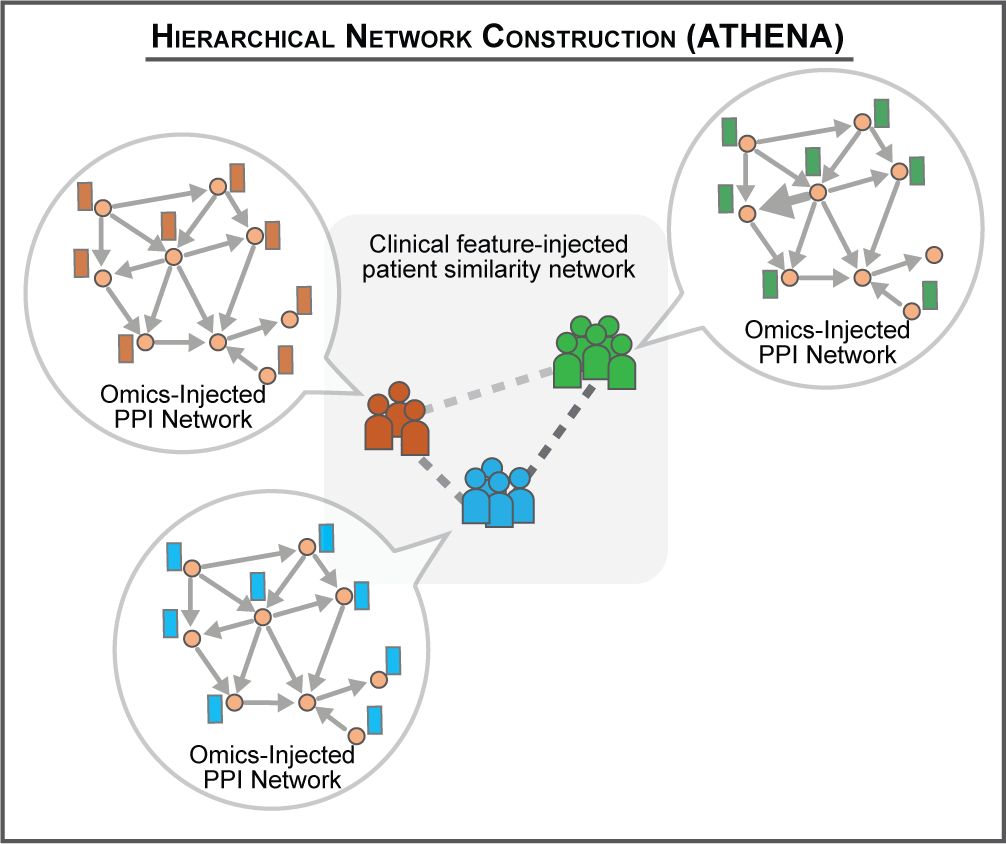}
    \caption{Transcriptomics-injected patient-specific PPI networks are embedded within their respective clinical features nodes in a patient similarity network.}
    \label{fig: Construction}
\end{figure}

To leverage the benefits of molecular signatures in phenotypic diagnosis, investigations in omics-based classification techniques have emerged for stratifying disease risk and elucidating underlying drivers of pathogenesis. Traditional approaches often begin with differential expression analysis or penalized regression to identify a subset of features that are correlative of a diagnosis \cite{Libby}. Supervised classifiers such as support vector machines (SVMs), random forests, and elastic-net logistic regression have then been trained on the feature subset to differentiate and classify phenotypes \cite{Hansson, Maiwald}. Due to the inherent noisy and nonlinear nature of expression data, recent deep-learning methods employ autoencoders or multi-layer perceptrons to learn latent representation of molecular profiles to generate a more robust and effective differentiation \cite{Zhao}. However, many models treat each molecular profile as an unstructured vector and ignore the underlying mechanistic insights reflected in molecular interactions of cellular networks and pathway topologies. We address this challenge by applying graph deep learning models that have been shown to effectively integrate and identify these molecular interactions towards a specific phenotype. Briefly, by embedding molecules as nodes within biological networks and using graph convolutional networks (GCNs) and graph attention networks (GATs) to propagate molecular expression signals between edges, we achieved a representation reflecting both the omics profile and its underlying pathogenesis of the clinical sub-phenotypes \cite{Way, Taroni, Wang, Pfeifer, Schulte-Sasse}. 

Recent advancements in graph-supported deep learning has expanded our understanding of cellular pathways and networks with respect to genotypes and function. However, effective translation of this field faces major limitations: first, these workflows are limited to a single homogeneous graph and its subgraph relevant to disease, but cannot model individual patient variabilities and understand the heterogeneity of a cohort. Furthermore, these approaches focus solely on molecular information without considering cohort-wide population phenotypes such as shared comorbidities and demographic similarities, which are crucial for developing risk stratification strategies \cite{ZhouGNN}. Finally, there are challenges in model design to characterize subclasses of patient pathogenesis based on their molecular data, a key component for our understanding of major diseases, e.g., subclinical atherosclerosis \cite{Rajadhyaksha}.

To address these specific limitations, we propose a novel graph-based approach that models two essential aspects of a patient: clinical features and molecular data. We first create a patient similarity network based on their clinical features, thereby modelling cohort-level patterns and discovering clinically relevant subpopulations. By representing each patient as a node and weighting each edge according to their clinical feature-based similarities, the network consists of patient subgroups exhibiting similar conditions \cite{Pai, Tang}. Molecular interaction networks encode mechanistic insights by treating proteins as nodes and their interactions as edges, enabling the identification of disease-relevant subnetworks \cite{Xie}. We demonstrate that jointly embedding patient-specific molecular networks and cohort-level similarity networks into a unified hierarchical graph has shown a holistic representation of disease phenotypes, which improves the accuracy of subclinical atherosclerosis classification and facilitates robust patient subtyping. By capturing both individual perturbations and cohort-level population patterns, our framework delivers interpretable mechanistic insights to better understand disease pathogenesis and to aid personalized intervention strategies.

In a nutshell, we introduce a hierarchical integration framework and graph‐representation learning pipeline that jointly leverages patient‐specific molecular networks and cohort‐level similarity graphs. We evaluate our approach on a cardiovascular case study utilizing a cohort from Sanchez-Cabo et al. \cite{Sanchez} and from Steenman et al. \cite{Steenman}, as well as demonstrate that a hierarchical representation for patient stratification outperforms previous graph deep learning approaches for classification. Importantly, we demonstrate the capacity of our approach pertaining to disease subtype identification by clustering the learned protein interactions of patient-specific molecular sub-networks. This approach is novel as it combines the endogenous molecular features unique to each patient with that of clinical features reflecting cohort-level patterns, resulting in early risk stratification and interpretable subtype identification to assist targeted intervention strategies.

\begin{table*}[]
\begin{tabular}{|l|cccccc|}
\hline
\multicolumn{1}{|c|}{\multirow{3}{*}{\textbf{Graph Type}}} & \multicolumn{6}{c|}{\textbf{Datasets}}                                                                                                                                                                                                \\ \cline{2-7} 
\multicolumn{1}{|c|}{}                                     & \multicolumn{3}{c|}{\textbf{Sanchez-Cabo et al. \cite{Sanchez}}}                                                                         & \multicolumn{3}{c|}{\textbf{Steenman et al. \cite{Steenman}}}                                                        \\ \cline{2-7} 
\multicolumn{1}{|c|}{}                                     & \multicolumn{1}{c|}{\textbf{\# Nodes}} & \multicolumn{1}{c|}{\textbf{\# Edges}} & \multicolumn{1}{c|}{\textbf{Avg. Degree}} & \multicolumn{1}{c|}{\textbf{\# Nodes}} & \multicolumn{1}{c|}{\textbf{\# Edges}} & \textbf{Avg. Degree} \\ \hline
\textbf{Patient Similarity Network}                        & \multicolumn{1}{c|}{391 Patients}      & \multicolumn{1}{c|}{19,550 Edges}      & \multicolumn{1}{c|}{50 Edges}             & \multicolumn{1}{c|}{104 Patients}      & \multicolumn{1}{c|}{2080 Edges}        & 20 Edges             \\ \hline
\textbf{STRING PPI Network}                                & \multicolumn{1}{c|}{645 Proteins}      & \multicolumn{1}{c|}{5,327 Edges}       & \multicolumn{1}{c|}{8.26 Edges}           & \multicolumn{1}{c|}{327 Proteins}      & \multicolumn{1}{c|}{3229 Edges}        & 9.87 Edges           \\ \hline
\textbf{Hierarchical Network (ATHENA)}                     & \multicolumn{1}{c|}{252,195 Nodes}     & \multicolumn{1}{c|}{2,102,234 Edges}   & \multicolumn{1}{c|}{8.34 Edges}           & \multicolumn{1}{c|}{34,008 Nodes}      & \multicolumn{1}{c|}{337,896 Edges}     & 9.93 Edges           \\ \hline
\end{tabular}
\caption{Characteristics of Network Topologies Supporting Graph Construction Strategies. Illustration of node count, edge count, and average degree for the three graph types, including a graph representing clinical features of patients in a cohort setting (patient similarity network), a graph representing a patient-specific STRING Protein-Protein Interaction (PPI) Network cross-referenced with individualized transcriptomics features, and a hierarchical network (i.e., ATHENA) that integrates the cohort-context patient similarity graph with each patients’ respective PPI network. These network characteristics provide context for graph complexity and connectivity in downstream graph-based analyses.}
\label{tab:Characteristics}
\end{table*}

\section{Problem Statement}
In this work, we study the problem of personalized classification of subclinical atherosclerosis by developing a hierarchical graph neural network framework to leverage two clinical aspects of a patient: clinical features (e.g., demographics, comorbidities) and molecular data (e.g., filtered omics features mapped onto patient-specific PPI subgraphs). Our aim is to enhance precision cardiovascular diagnostics and to identify disease subtypes by integrating molecular interaction signatures of each patient with clinical features that reflect cohort-level behaviors in our predictive model.

Integrating these dual clinical aspects presents significant challenges:
\begin{enumerate}
    \item Heterogeneous Data Alignment \& Graph Construction: Clinical features and transcriptomics are distinct data types and differ widely in scale, information, and modality, therefore challenging to directly align them into a unified representation. 
    \item Integrated Modality Learning: Optimization of patient-specific molecular fingerprints that reflect individual omics data while enforcing consistency with cohort-wide patterns demands a sensitive balance of cross-modality representation objectives. 
    \item Patient Subtype Identification: Patient subtyping often considers only clinical feature similarity, overlooking shared pathogenic interactions. Incorporating shared disease sub-networks into stratification enables mechanistically-informed subtype discovery and personalized intervention strategies.
\end{enumerate}

To address these challenges, we introduce ATHENA: \textbf{A}therosclerosis \textbf{T}hrough \textbf{H}ierarchical \textbf{E}xplainable \textbf{N}etwork \textbf{A}nalysis, which creates a hierarchical network representation that harmonizes patient-specific molecular interactions within the context of the disease cohort. We show that this integration method not only boosts subclinical atherosclerosis classification performance but also enables mechanistically informed subtype discovery through XAI-driven subnetwork clustering.

\section{Methods}


\subsection{Datasets and Preprocessing}
We employed a transcriptomics dataset from the Progression of Early Subclinical Atherosclerosis (PESA) study \cite{Sanchez}. Sanchez-Cabo et al. reported whole blood transcriptomics data from 391 patients. In this study, subclinical atherosclerosis burden was assessed using vascular ultrasound and coronary artery calcification (CAC) scores, enabling the identification of subclinical phenotypic associations, in which molecular profiles across different stages of disease development were measured using RNA sequencing (RNA-seq) of polyadenylated RNA extracted from peripheral blood samples. From these CAC scores, a clinician assigned each patient one of four PESA scores from the ultrasound study: None, Generalized, Intermediate, and Focal. For our analysis, we utilized the plasma transcriptomics assay data, comprising measurements for 12,061 genes per patient, and patient clinical features. To validate our analysis, we also employed another transcriptomics dataset with 104 patients from Steenman et al. \cite{Steenman}, which includes RNA sequencing data from human arterial tissue samples collected post-mortem. 

For preprocessing transcriptomics, we utilized a variance threshold of 0.1, followed by an iterative one-way analysis-of-variance (ANOVA) with a Bonferroni-adjusted p-value threshold of 0.01 to filter out transcriptomics features with minimal variability across patients. The selected transcriptomics features establish a baseline for interpreting differential gene expression across the four PESA scores. Next, we cross-referenced the genes from the PESA study to the STRING PPI database \cite{Szklarczyk} and applied a high-confidence threshold of 0.7 to remove statistically insignificant associations. Our final transcriptomics dataset consisted of 645 protein-coding genes with 5,327 corresponding interactions from STRING. Using the same preprocessing steps for our second dataset, we obtained a final transcriptomics dataset comprising 327 protein-coding genes and 3,229 corresponding interactions from STRING.

\subsection{Hierarchical Graph Construction}

Patient-specific PPI networks are embedded within their respective patient nodes, which are interconnected to form an overarching patient similarity network. For the STRING PPI networks, protein-coding genes were aligned to their corresponding downstream proteins. The transcriptomics (omics) matrix was then cross-referenced with the PPI network, with the resulting omics data injected into their respective protein nodes to personalize the network for each patient (\autoref{fig: Construction}). 

\begin{figure*}[th]
    \includegraphics[width=\textwidth, keepaspectratio]{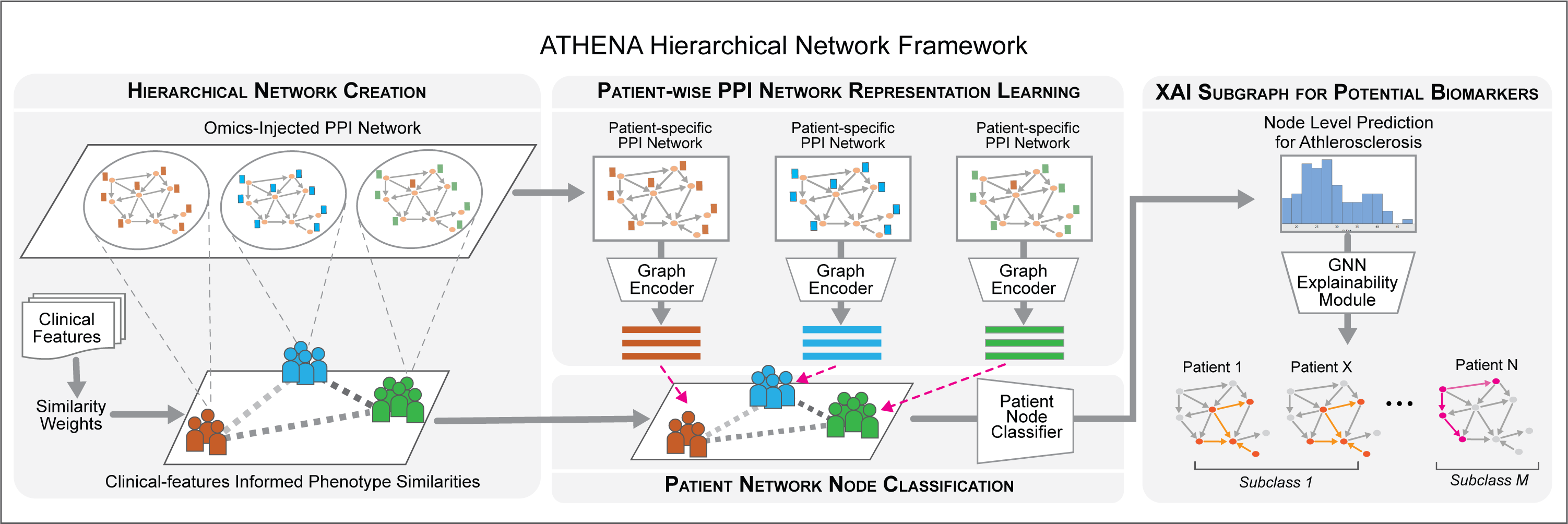}
    \caption{Pipeline for multi-modal hierarchical graph deep learning for patient classification. Graph encoders are applied to each patient's PPI network to generate feature representations that capture both structural and omics-informed embeddings. These embeddings serve as inputs to every patient node in the patient similarity network, which is then processed using a GNN to classify atherosclerotic burden. An explainable AI (XAI) module is linked to the GNN to extract subgraphs from the patient-specific PPI networks, highlighting potential patient-level biomarkers contributing to the predictions.}
    \label{fig: Architecture}
\end{figure*}

Let $\mathcal{G}_i = (\mathcal{V}_i, \mathcal{E}_i)$ denote the personalized protein–protein interaction (PPI) network for patient $i$, where $\mathcal{V}_i$ represents the set of protein nodes and $\mathcal{E}_i \subseteq \mathcal{V}_i \times \mathcal{V}_i$ the set of interactions. Each $\mathcal{G}_i$ is derived from the reference STRING PPI network by aligning expressed protein-coding genes from the transcriptomics matrix to their corresponding protein. Let $X_i \in \mathbb{R}^{|\mathcal{V}_i| \times d}$ represent the patient-specific omics feature matrix, where each row corresponds to a protein node enriched with expression values. This embedding personalizes the PPI graph $\mathcal{G}_i$ for patient $i$.

To construct the patient similarity network, we compute pairwise similarities between patients based on their clinical feature vectors. Let $\mathbf{c}_i \in \mathbb{R}^m$ denote the clinical feature vector for patient $i$, where $m$ is the number of features. The similarity between patients $i$ and $j$ is defined using a Gaussian (RBF) kernel:

\[
S_{ij} = \exp\left(-\frac{\|\mathbf{c}_i - \mathbf{c}_j\|^2}{2\sigma^2}\right),
\]

\noindent where $\sigma$ is a bandwidth parameter controlling the width of the kernel. A $k$-nearest neighbors (kNN) strategy is applied to retain only the top-$k$ most similar patients for each node, resulting in a sparse similarity matrix $S$. For the dataset from Sanchez et al. \cite{Sanchez}, we selected $k=50$, and for Steenman et al. \cite{Steenman}, $k=20$.

These similarity scores are used as weighted edges to construct the patient similarity graph $\mathcal{H} = (\mathcal{U}, W)$, where each node $u_i \in \mathcal{U}$ represents a patient and $W_{ij} = S_{ij}$ encodes the edge weight. This graph-based representation captures cohort-wide relationships for downstream network analysis.

\subsection{Model Architecture and Subtype Characterization}

Our workflow for our hierarchical graph deep learning model (ATHENA) is illustrated in \autoref{fig: Architecture}. Using the clinical features and filtered transcriptomics data, we constructed a patient similarity network to identify initial patient subgroups, with each patient having their own PPI network. The patient similarity network is constructed from each patient’s clinical features, where each node represents a patient and each edge represents the similarity between the top $k$ neighboring patients, capturing cohort-level patterns and identifying clinically meaningful subgroups. Further, each patient node contains a patient-specific protein-protein interaction (PPI) network constructed from integrating omics data with the PPI network from STRING. This dual approach integrates clinical and molecular information, enabling a comprehensive characterization of patient-specific disease mechanisms and facilitating the identification of phenotype-driven molecular signatures.

Each patient node $u_i$ in the similarity graph $\mathcal{H} = (\mathcal{U}, W)$ is associated with a personalized protein–protein interaction network $\mathcal{G}_i = (\mathcal{V}_i, \mathcal{E}_i, X_i)$, where $X_i \in \mathbb{R}^{|\mathcal{V}_i| \times d}$ represents omics-informed node features. These personalized molecular graphs integrate STRING-derived interaction topology with transcriptomics data, enabling patient-specific modeling of molecular mechanisms.

To create a representation for the entire molecular interaction graph, we apply a graph neural network (GNN) encoder \( \phi_{\text{PPI}} \) to each patient-specific PPI graph \( \mathcal{G}_i = (\mathcal{V}_i, \mathcal{E}_i, X_i) \), where \( \mathcal{V}_i \) and \( \mathcal{E}_i \) denote the set of protein nodes and interactions, respectively, and \( X_i \) contains omics-informed node features. The GNN encoder computes hidden node representations through \( L \) layers of message passing, starting from initial node features \( \mathbf{h}_v^{(0)} = \mathbf{x}_v \), where \( \mathbf{x}_v \) is the omics feature vector for protein node \( v \). At each layer \( \ell = 1, \dots, L \), the hidden representation of node \( v \in \mathcal{V}_i \) is updated as:

\begin{equation}
\mathbf{h}_v^{(\ell)} = \sigma \left( \mathbf{W}^{(\ell)} \cdot \text{AGGREGATE}^{(\ell)} \left( \left\{ \mathbf{h}_u^{(\ell-1)} : u \in \mathcal{N}(v) \cup \{v\} \right\} \right) \right),
\end{equation}

\noindent where \( \mathcal{N}(v) \) denotes the set of neighbors of \( v \), \( \mathbf{W}^{(\ell)} \) is a trainable weight matrix, \( \sigma \) is a non-linear activation function (e.g., ReLU), and \( \text{AGGREGATE}^{(\ell)} \) is a permutation-invariant function such as mean, sum, or attention. After the final layer, we obtain the set of node representations \( \{ \mathbf{h}_v^{(L)} \}_{v \in \mathcal{V}_i} \), which are aggregated using a READOUT function (e.g., max pooling) to produce a graph-level embedding:

\begin{equation}
\mathbf{z}_i = \phi_{\text{PPI}}(\mathcal{G}_i) = \text{READOUT}\left( \left\{ \mathbf{h}_v^{(L)} : v \in \mathcal{V}_i \right\} \right).
\end{equation}

To facilitate cross-modality learning and ensure that molecular signatures align with cohort-wide features, the patient-specific molecular embeddings \( \mathbf{z}_i \in \mathbb{R}^d \), obtained from the graph-level encoder \( \phi_{\text{PPI}}(\mathcal{G}_i) \), are used as node features in the patient similarity graph \( \mathcal{H} = (\mathcal{U}, W) \). Each patient node \( u_i \in \mathcal{U} \) is initialized with feature vector \( \mathbf{h}_i^{(0)} = \mathbf{z}_i \).

A graph neural network \( \phi_{\text{cohort}} \) is then applied to this cohort-level graph to propagate and refine patient representations with respect to their clinical similarity topology. At each layer \( \ell = 1, \dots, L' \), the hidden representation of patient node \( u_i \) is updated as:

\begin{equation}
\mathbf{h}_i^{(\ell)} = \sigma \left( \mathbf{W}^{(\ell)} \cdot \text{AGGREGATE}^{(\ell)} \left( \left\{ \mathbf{h}_j^{(\ell-1)} : j \in \mathcal{N}(i) \cup \{i\} \right\} \right) \right),
\end{equation}

\noindent where \( \mathcal{N}(i) \) denotes the set of clinically similar neighbors of patient \( i \), \( \mathbf{W}^{(\ell)} \) is a trainable weight matrix, and \( \sigma \) is a non-linear activation function (e.g., ReLU).

The final patient representation \( \mathbf{h}_i^{(L')} \) encodes information from both the patient’s molecular graph and the patient's phenotypic neighborhood in the clinical similarity graph. These representations are then used for downstream prediction of atherosclerotic burden:

\begin{equation}
\hat{y}_i = \phi_{\text{cohort}}(\mathbf{h}_i^{(L')}),
\end{equation}

\noindent This hierarchical learning framework allows the model to jointly reason over individual molecular signatures with cohort-wide clinical structure.


Finally, an explainable AI (XAI) module extracts subgraphs from the ATHENA’s predictions, revealing patient-specific biomarkers most influential for a patient's atherosclerotic burden. Specifically, we employ GNNExplainer \cite{Ying} to extract a subgraph \( \mathcal{G}_i^{\text{XAI}} \subseteq \mathcal{G}_i \) that maximizes the mutual information between the explanation and the prediction \( \hat{y}_i \). Formally, GNNExplainer optimizes the following objective:

\begin{equation}
\max_{\mathcal{G}_i^{\text{XAI}},\, X_i^{\text{XAI}}} \; \mathbb{MI} \left( \hat{y}_i ; \mathcal{G}_i^{\text{XAI}}, X_i^{\text{XAI}} \right),
\end{equation}

\noindent where \( \mathbb{MI}(\cdot;\cdot) \) denotes mutual information. This process reveals the patient-specific molecular interaction profiles most influential for the atherosclerotic burden prediction, thus enabling biomarker discovery.

The resulting molecular explanations \( \{ \mathcal{G}_i^{\text{XAI}} \} \) are then used for clustering patients through t-SNE. This approach facilitates the identification of atherosclerosis clusters by leveraging interpretable, learned interaction patterns within each patient's graph representation. Patients are clustered utilizing these learned patient-specific molecular interaction profiles, highlighting an innovative approach in cluster identification within each atherosclerosis subtype.

\section{Results}
\subsection{Evaluation Metrics}

For model training, we utilized the Adam optimizer with a learning rate of 0.01 and a weight decay of $1e^{-6}$. We also set the number of epochs to 100. We employed 5-fold nested cross validation, in which the train set is split into train/validation for grid search optimization. Details of the grid search parameter space are shown in Supplementary Table 3 and Supplementary Table 4.

For model evaluations, we employed area under the receiver operator curve (AUC) and F1 score \cite{Rainio}. For benchmarking and in our ablation studies, both metrics were calculated across all phenotypes. For evaluating our Hierarchical Network Classification model on specific atherosclerosis phenotypes, only AUC was calculated. In all evaluations, both metrics were calculated across 5-fold stratified cross validation.

\subsection{Classification and Profiling Performance}

\begin{table}[]
\centering
\footnotesize  
\setlength{\tabcolsep}{5pt}  
\renewcommand{\arraystretch}{1.1}  
{
\begin{tabular}{|lcccc|}
\hline
\multicolumn{1}{|c|}{}                                                    & \multicolumn{4}{c|}{\textbf{Datasets}}                                                                                                                                                                                        \\ \cline{2-5} 
\multicolumn{1}{|c|}{}                                                    & \multicolumn{2}{c|}{Sanchez-Cabo et al. \cite{Sanchez}}                                                                              & \multicolumn{2}{c|}{Steenman et al. \cite{Steenman}}                                                             \\ \cline{2-5} 
\multicolumn{1}{|c|}{\multirow{-3}{*}{\textbf{Models}}}                    & \multicolumn{1}{c|}{\textbf{AUC}}                          & \multicolumn{1}{c|}{\textbf{Macro F1}}                           & \multicolumn{1}{c|}{\textbf{AUC}}                          & \textbf{Macro F1}                           \\ \hline
\multicolumn{5}{|l|}{\cellcolor[HTML]{D9D9D9}\textbf{Linear Modeling Approaches}}                                                                                                                                                                                                                        \\ \hline
\multicolumn{1}{|l|}{Lasso}                                               & \multicolumn{1}{c|}{0.681 ± 0.04}                          & \multicolumn{1}{c|}{0.553 ± 0.02}                          & \multicolumn{1}{c|}{0.753 ± 0.009}                         & 0.743 ± 0.013                         \\ \hline
\multicolumn{1}{|l|}{Elastic Net}                                         & \multicolumn{1}{c|}{0.681 ± 0.04}                          & \multicolumn{1}{c|}{0.555 ± 0.02}                          & \multicolumn{1}{c|}{0.774 ± 0.012}                         & 0.757 ± 0.014                         \\ \hline
\multicolumn{1}{|l|}{SVC}                                                 & \multicolumn{1}{c|}{0.684 ± 0.02}                          & \multicolumn{1}{c|}{0.612 ± 0.02}                          & \multicolumn{1}{c|}{0.785 ± 0.008}                         & 0.778 ± 0.011                         \\ \hline
\multicolumn{5}{|l|}{\cellcolor[HTML]{D9D9D9}\textbf{Deep Learning Approaches}}                                                                                                                                                                                                                          \\ \hline
\multicolumn{1}{|l|}{MLP (3-layer)}                                       & \multicolumn{1}{c|}{\cellcolor[HTML]{FFFFFF}0.712 ± 0.05}  & \multicolumn{1}{c|}{\cellcolor[HTML]{FFFFFF}0.533 ± 0.06}  & \multicolumn{1}{c|}{0.805 ± 0.007}                         & 0.786 ± 0.012                         \\ \hline
\multicolumn{1}{|l|}{MLP (5 layer)}                                       & \multicolumn{1}{c|}{\cellcolor[HTML]{FFFFFF}0.708 ± 0.03}  & \multicolumn{1}{c|}{\cellcolor[HTML]{FFFFFF}0.525 ± 0.03}  & \multicolumn{1}{c|}{0.827 ± 0.010}                         & 0.808 ± 0.009                         \\ \hline
\multicolumn{5}{|l|}{\cellcolor[HTML]{C0C0C0}\textbf{Graph Deep Learning (Patient Similarity)}}                                                                                                                                                                                                          \\ \hline
\multicolumn{1}{|l|}{GCN} & \multicolumn{1}{c|}{0.747 ± 0.042}                         & \multicolumn{1}{c|}{0.684 ± 0.030}                         & \multicolumn{1}{c|}{0.826 ± 0.011}                         & 0.806 ± 0.010                         \\ \hline
\multicolumn{1}{|l|}{GAT}    & \multicolumn{1}{c|}{0.753 ± 0.043}                         & \multicolumn{1}{c|}{0.676 ± 0.031}                         & \multicolumn{1}{c|}{0.823 ± 0.014}                         & 0.817 ± 0.013                         \\ \hline
\multicolumn{1}{|l|}{GraphSage}                    & \multicolumn{1}{c|}{0.748 ± 0.045}                         & \multicolumn{1}{c|}{0.680 ± 0.029}                         & \multicolumn{1}{c|}{0.817 ± 0.013}                         & 0.799 ± 0.012                         \\ \hline
\multicolumn{5}{|l|}{\cellcolor[HTML]{CCCCCC}\textbf{Hierarchical Network Classification (ATHENA)}}                                                                                                                                                                                                      \\ \hline
\multicolumn{1}{|l|}{GCN}   & \multicolumn{1}{c|}{\cellcolor[HTML]{B6D7A8}0.812 ± 0.062} & \multicolumn{1}{c|}{\cellcolor[HTML]{B6D7A8}0.725 ± 0.072} & \multicolumn{1}{c|}{\cellcolor[HTML]{B6D7A8}0.837 ± 0.013} & \cellcolor[HTML]{B6D7A8}0.832 ± 0.011 \\ \hline
\multicolumn{1}{|l|}{GAT}    & \multicolumn{1}{c|}{\cellcolor[HTML]{B6D7A8}0.808 ± 0.058} & \multicolumn{1}{c|}{\cellcolor[HTML]{B6D7A8}0.718 ± 0.070} & \multicolumn{1}{c|}{\cellcolor[HTML]{B6D7A8}0.843 ± 0.009} & \cellcolor[HTML]{B6D7A8}0.839 ± 0.010 \\ \hline
\multicolumn{1}{|l|}{GraphSage}                    & \multicolumn{1}{c|}{\cellcolor[HTML]{B6D7A8}0.805 ± 0.060} & \multicolumn{1}{c|}{\cellcolor[HTML]{B6D7A8}0.721 ± 0.068} & \multicolumn{1}{c|}{\cellcolor[HTML]{B6D7A8}0.839 ± 0.011} & \cellcolor[HTML]{B6D7A8}0.836 ± 0.014 \\ \hline
\end{tabular}}
\caption{Performance Metrics of Omics-based Patient Classification Models. Comparison of classification performance across multiple models and feature combinations using metrics tested. Results are presented as mean ± standard deviation over 5 fold stratified cross validation. Models incorporating the Hierarchical Network representation (ATHENA) to generate graph topology achieved the highest performance across all metrics. }
\label{tab:performance}
\end{table}

In \autoref{tab:performance}, we examined the performance of our hierarchical network representation (ATHENA) against linear modeling (Lasso, Elastic-Net with SAGA optimizer and an L1 ratio of 0.5, and Support Vector Classifier (SVC) \cite{Khan}), deep learning (3-layer and 5-layer multi-layer perceptron (MLP) \cite{Khan}), and graph deep learning (Graph Convolutional Network \cite{Kipf}, Graph Attention Network \cite{Velivckovic}, and GraphSage \cite{Hamilton}) methods. The mean and standard deviation of AUC and F1 score for all methods are summarized (Table 2), with the best performing models highlighted in green. In general, graph-based approaches outperform non-graph methods, and integrating molecular interaction topology alongside clinical features leads to added performance gains. Notably, ATHENA’s hierarchical network representation contributed to the uniform performance gains across all graph deep learning methods evaluated.

\begin{figure*}
    \includegraphics[width=0.8\textwidth, keepaspectratio]{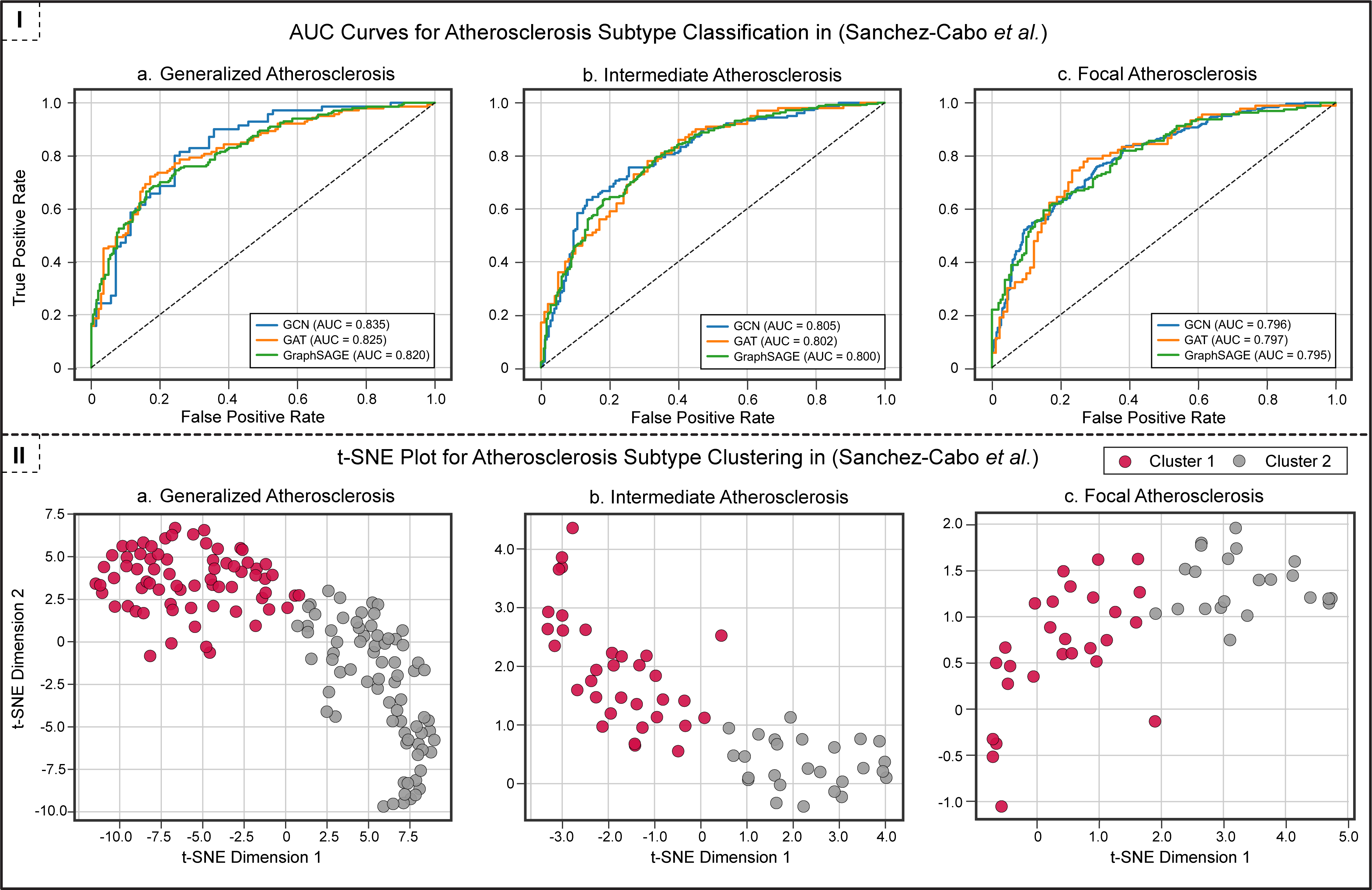}
    \caption{Model performance against individual subtypes of atherosclerosis for Sanchez-Cabo et al. \cite{Sanchez}. Upper panels show the consistent performance of ATHENA across all 3 subtypes in their mean AUC curves. Lower panels show the t-SNE-based separation of 2 distinct clusters in each subtype of atherosclerosis. Each dot represents one patient. For generalized atherosclerosis, cluster 1 has 27 patients and cluster 2 has 22 patients. For intermediate atherosclerosis, cluster 1 has 34 patients and cluster 2 has 27 patients. For focal atherosclerosis, cluster 1 has 75 patients and cluster 2 has 71 patients.}
    \label{fig: SanchezPerformance}
\end{figure*}

Next, we characterize the Hierarchical Network Classification model (ATHENA) in predicting specific atherosclerosis subtypes. The top row of \autoref{fig: SanchezPerformance}, presents the mean ROC curves for ATHENA in predicting Generalized (Figure 3Ia), Intermediate (Figure 3Ib), and Focal atherosclerosis (Figure 3Ic). AUC values ranged from 0.820 to 0.835 for Generalized, 0.800 to 0.805 for Intermediate, and 0.795-0.796 for Focal, indicating consistent predictive power across different subtypes. Further, panel I of \autoref{fig: SteenmanPerformance} presents the mean ROC curve for ATHENA. AUC values ranged between 0.835 and 0.846, indicating consistent performance across the three ATHENA architectures. 

\subsection{Ablation Studies}
To demonstrate the effectiveness of ATHENA, we scrutinized three model configurations: (1) patient similarity network only with transcriptomic markers and clinical features, (2) PPI network only with transcriptomics markers, and (3) ATHENA. \autoref{tab:ablation} summarizes the mean and standard deviation of AUC and F1 score for each configuration. Using only a patient similarity network representation resulted in the lowest performance, whereas relying solely on the PPI network representation produced slightly better results. The integration of both types of networks into a unified graph data representation achieved the highest overall performance across all metrics, underscoring the notion that a hierarchical network integration is the optimal method. Compared to patient similarity networks alone, ATHENA has had average AUC improvements of 0.057 to 0.065 and F1 score improvements of 0.041 to 0.042 for the cohort from Sanchez-Cabo et al. \cite{Sanchez}. For the cohort from Steenman et al. \cite{Steenman}, ATHENA has had average AUC improvements of 0.011 to 0.025 and F1 score improvements of 0.022 to 0.037. These results highlight that it is the synergistic representation provided by combining the two networks, not just the deep learning approach, that drives performance gains.

\subsection{Discovery of Pathogenesis Clusters}

ATHENA identified two clusters of patients that share similar molecular interaction profiles by utilizing the embedded XAI subnetwork for clustering. The atherosclerosis subtypes (e.g., Generalized, Intermediate, and Focal) were originally determined through clinician-validated medical imaging based on plaque buildup severity. However, the molecular interaction clustering revealed by ATHENA suggests two clusters of patients within each imaging-defined subtype, possibly reflecting different stages of disease progression. Row II of \autoref{fig: SanchezPerformance} presents t-SNE visualizations of patient embeddings derived from ATHENA for Generalized (Figure 3IIa), Intermediate (Figure 3IIb), and Focal atherosclerosis (Figure 3IIc), revealing clear separation of patients into two clusters within each atherosclerosis subtype.

\begin{table}[h]
\centering
\footnotesize  
\setlength{\tabcolsep}{5pt}  
\renewcommand{\arraystretch}{1.1}  
\begin{tabular}{|lcccc|}
\hline
\multicolumn{1}{|c|}{}                                                  & \multicolumn{4}{c|}{\textbf{Datasets}}                                                                                                                                                                                        \\ \cline{2-5} 
\multicolumn{1}{|c|}{}                                                  & \multicolumn{2}{c|}{Sanchez-Cabo et al. \cite{Sanchez}}                                                                              & \multicolumn{2}{c|}{Steenman et al. \cite{Steenman}}                                                             \\ \cline{2-5} 
\multicolumn{1}{|c|}{\multirow{-3}{*}{\textbf{Models}}}                  & \multicolumn{1}{c|}{\textbf{AUC}}                          & \multicolumn{1}{c|}{\textbf{Macro F1}}                           & \multicolumn{1}{c|}{\textbf{AUC}}                          & \textbf{Macro F1}                           \\ \hline
\multicolumn{5}{|l|}{\cellcolor[HTML]{D9D9D9}\textbf{Node Classification (Similarity Network)}}                                                                                                                                                                                                        \\ \hline
\multicolumn{1}{|l|}{GCN} & \multicolumn{1}{c|}{0.747 ± 0.042}                         & \multicolumn{1}{c|}{0.684 ± 0.030}                         & \multicolumn{1}{c|}{0.826 ± 0.011}                         & 0.806 ± 0.010                         \\ \hline
\multicolumn{1}{|l|}{GAT}  & \multicolumn{1}{c|}{0.753 ± 0.043}                         & \multicolumn{1}{c|}{0.676 ± 0.031}                         & \multicolumn{1}{c|}{0.823 ± 0.014}                         & 0.817 ± 0.013                         \\ \hline
\multicolumn{1}{|l|}{GraphSage}                  & \multicolumn{1}{c|}{0.748 ± 0.045}                         & \multicolumn{1}{c|}{0.680 ± 0.029}                         & \multicolumn{1}{c|}{0.817 ± 0.013}                         & 0.799 ± 0.012                         \\ \hline
\multicolumn{5}{|l|}{\cellcolor[HTML]{D9D9D9}\textbf{Graph Classification (PPI Network)}}                                                                                                                                                                                                              \\ \hline
\multicolumn{1}{|l|}{GCN} & \multicolumn{1}{c|}{0.760 ± 0.05}                          & \multicolumn{1}{c|}{0.705 ± 0.06}                          & \multicolumn{1}{c|}{0.803 ± 0.010}                          & 0.787 ± 0.010                          \\ \hline
\multicolumn{1}{|l|}{GAT}  & \multicolumn{1}{c|}{0.765 ± 0.05}                          & \multicolumn{1}{c|}{0.708 ± 0.06}                          & \multicolumn{1}{c|}{0.812 ± 0.012}                          & 0.792 ± 0.011                          \\ \hline
\multicolumn{1}{|l|}{GraphSage}                  & \multicolumn{1}{c|}{0.768 ± 0.05}                          & \multicolumn{1}{c|}{0.709 ± 0.06}                          & \multicolumn{1}{c|}{0.798 ± 0.011}                          & 0.778 ± 0.012                          \\ \hline
\multicolumn{5}{|l|}{\cellcolor[HTML]{CCCCCC}\textbf{Hierarchical Network Classification (ATHENA)}}                                                                                                                                                                                                    \\ \hline
\multicolumn{1}{|l|}{GCN} & \multicolumn{1}{c|}{\cellcolor[HTML]{B6D7A8}0.812 ± 0.062} & \multicolumn{1}{c|}{\cellcolor[HTML]{B6D7A8}0.725 ± 0.072} & \multicolumn{1}{c|}{\cellcolor[HTML]{B6D7A8}0.837 ± 0.013} & \cellcolor[HTML]{B6D7A8}0.832 ± 0.011 \\ \hline
\multicolumn{1}{|l|}{GAT}  & \multicolumn{1}{c|}{\cellcolor[HTML]{B6D7A8}0.808 ± 0.058} & \multicolumn{1}{c|}{\cellcolor[HTML]{B6D7A8}0.718 ± 0.070} & \multicolumn{1}{c|}{\cellcolor[HTML]{B6D7A8}0.843 ± 0.009} & \cellcolor[HTML]{B6D7A8}0.839 ± 0.010 \\ \hline
\multicolumn{1}{|l|}{GraphSage}                  & \multicolumn{1}{c|}{\cellcolor[HTML]{B6D7A8}0.805 ± 0.060} & \multicolumn{1}{c|}{\cellcolor[HTML]{B6D7A8}0.721 ± 0.068} & \multicolumn{1}{c|}{\cellcolor[HTML]{B6D7A8}0.839 ± 0.011} & \cellcolor[HTML]{B6D7A8}0.836 ± 0.014 \\ \hline
\end{tabular}
\caption{Ablation Results for ATHENA. Various ablation configurations of ATHENA on various classification metrics. The configurations include the Patient Similarity Network alone, the Protein-Protein Interaction (PPI) Network alone, and ATHENA. Integrating hierarchical structures and clinical features significantly improves performance, as demonstrated by ATHENA, achieving the highest scores across all metrics.}
\label{tab:ablation}
\end{table}

Our molecular clustering analysis shows that patients within the same imaging-defined subtype (None, Generalized, Intermediate, Focal) exhibit distinct molecular signatures, revealing opportunities for more granular patient stratification beyond categorization of anatomical features derived from imaging. Specifically, ATHENA identified two distinct molecular clusters within each imaging-defined subtype (Figure 3, Panel II), highlighting that patients with identical imaging presentations may be driven by fundamentally different underlying mechanisms. Rather than discrete imaging categories, our molecular analysis suggests atherosclerosis exists on a continuum. 

In the generalized subtype, two distinct clusters capture the balance between endothelial inflammation and structural remodeling. Cluster 1 is characterized by dysfunctional and inflammatory pathways, which is highlighted by up-regulation of CDK5R1 and ICAM1, markers known to contribute to endothelial regulation in atherosclerotic patients \cite{bai2012cyclin, kitagawa2002involvement}. In contrast, Cluster 2 is characterized by structural and remodeling pathways, exhibiting higher TTLL5 which drives microtubule remodeling necessary for thickening of the arterial wall \cite{bompard2018csap}. Cluster 2 also exhibits an upregulation RASGRP3 expression, thereby favoring cytoskeletal adaptations over inflammation \cite{tang2014rasgrp3}.

Stories reviewed in the intermediate subtype show that Cluster 1 amplifies hallmark atherogenic processes within monocyte and inflammatory pathways with elevated ITGAM/CD11b expression, leading to thickening of the arterial wall as shown in previous atherosclerotic studies \cite{lou2024itgam}. Furthermore, MMP9 upregulation is seen with CD14 enhancing LPS/TLR4 signaling in infiltrating monocytes \cite{yamamoto2018impact, aggarwal2023platelets}. Metabolic sensors INSR and FOSL2 are also evident \cite{semenkovich2006insulin, yin2017activator}. In parallel, Cluster 2 exhibits regulatory and metabolic pathways, characterized by SMIM24, MAF, and FCER1A representing a subset of intermediate lesions engaging lipid-handling or anti-inflammatory programs to temper unchecked matrix remodeling \cite{tudzynski1980transformation, luo2014c}.

In the subtype of advanced focal atherosclerosis, Cluster 1 hosts oxidative-stress and inflammatory regulation \cite{guo2012overexpression}, pro-inflammatory cytokine release \cite{amini2004interplay}, and HSPB1 chaperones proteins under oxidative stress. Cluster 2 hosts fibrotic-matrix remodeling pathways, which drive fibrous-cap integrity \cite{uyama2017involvement}, ITGB7-dependent integrin adhesion, and CEMIP2-directed hyaluronan reorganization in the extracellular matrix \cite{yoshida2025new}, collectively stabilizing advanced plaques.

Overall, the two clusters within each of the three subtypes represent a molecular phenotype driven by inflammation pathways and a second molecular phenotype highlighted by regulation of structural proteins. 

\begin{figure}[h]
    \includegraphics[width=0.4\textwidth, keepaspectratio]{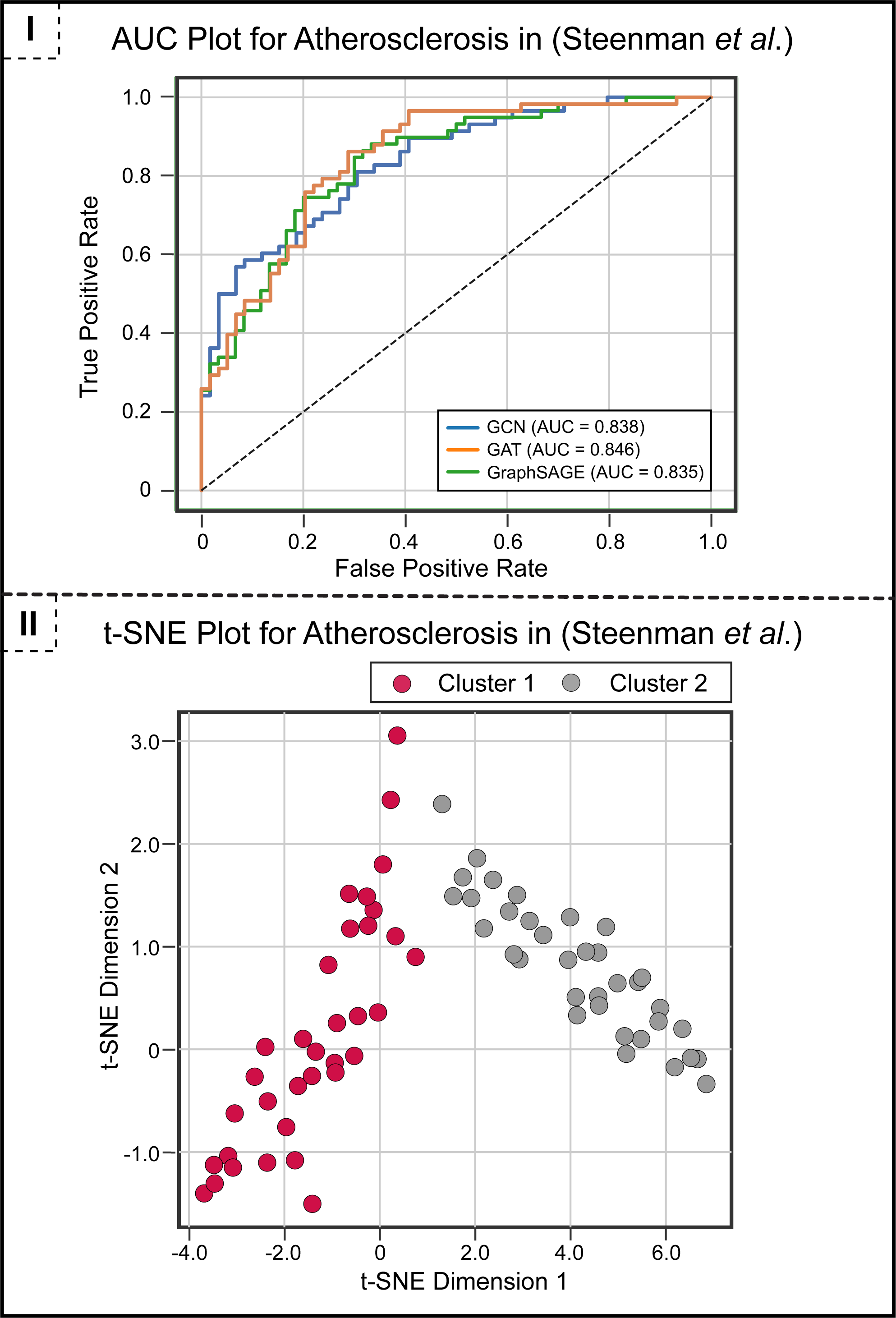}
    \caption{Model performance against cohort of atherosclerosis from Steenman et al. \cite{Steenman}. Panel I (top) shows consistent performance of ATHENA across the three architectures: GCN, GAT, and GraphSAGE. Panel II (bottom) shows the t-SNE-based separation of 2 distinct clusters in the cohort. Each dot represents 1 patient; cluster 1 has 34 patients; and cluster 2 has 36 patients.}
    \label{fig: SteenmanPerformance}
\end{figure}

We further investigate molecular heterogeneity in the cohort from Steenman et al. \cite{Steenman}. Panel II of \autoref{fig: SteenmanPerformance} presents clear separation between patients of the same clinical phenotype in two distinct clusters. Cluster 1 is characterized with upregulation of HOXA6 and HOXA5, which are involved in developmental patterning and vascular integrity \cite{Jing, Zaina}. In contrast, Cluster 2 exhibits elevated expression of SLC2A1 and SMS, genes associated with metabolic activation and cellular transport processes \cite{Wall, LiImpact}. These two clusters reveal molecular phenotypes that differ from those observed in the Sanchez et al. cohort \cite{Sanchez}, with distinctions in this cohort of one molecular phenotype primarily driven by regulation of structural proteins and a second one enriched in metabolic pathways.

\section{Discussion}
We present ATHENA, a novel hierarchical network representation that integrates both molecular interaction topology and clinical features. Linear modeling approaches and standard deep learning (multi-layer perceptron) treat molecular profiles as unstructured vectors; unfortunately, they overlook the mechanistic insights of pathogenesis embedded in molecular interaction networks and pathway topologies. In contrast, graph deep learning identifies and integrates these molecular interactions toward a specific phenotype. ATHENA extends this notion by combining patient clinical features and molecular data into a unified hierarchical graph, thereby creating a more comprehensive representation of disease phenotypes.  

We have demonstrated that ATHENA achieves superior performance in predicting atherosclerosis phenotypes compared to linear/standard deep learning models. This representation has consistent predictive power across different atherosclerosis subtypes from blood and arterial tissue transcriptome profiles. Our ablation studies highlight that the performance gains stem from the synergistic alignment of patient similarity networks and molecular interaction networks. Finally, our analysis of molecular heterogeneity reveals that our model can identify distinct patient clusters within each atherosclerosis subtype. 

While ATHENA demonstrates significant improvements in atherosclerosis subtype classification and patient clustering, several limitations warrant consideration for future development and clinical translation: 

\begin{enumerate}
    \item Our first dataset is restricted to whole blood transcriptomics. While blood-based biomarkers offer practical advantages for clinical implementation, they may not fully capture tissue-specific molecular mechanisms underlying atherosclerosis progression. With our second dataset, the tissue representation of molecular phenotypes has been addressed.
    \item Molecular characterizations of the patients highlight the regulatory involvement of structure, inflammation, and metabolic proteins. Although transcriptomics represent accurate changes in these molecules, integration of additional omics layers, such as proteomics or epigenomics, could provide a more comprehensive molecular portrait of disease pathogenesis.
    \item Our analysis captures molecular drivers for a patient cohort consisting of multiple stages of atherosclerosis (generalized, intermediate, and focal), representative of a cross-sectional evaluation of disease progression. ATHENA's performance is consistent across all stages as well as discovers unique molecular subgraphs and biomarkers for each stage. However, longitudinal studies tracking molecular and clinical changes over time would provide valuable insights into the dynamic nature of atherosclerosis development and enable precision in the prediction of disease trajectories.
    \item We acknowledge that comprehensive batch effect analysis would strengthen clinical applicability, where a commitment to extensive large cohort, multicenter clinical trials validations will be necessary.

\end{enumerate}

In summary, the superior performance of the hierarchical representation compared to isolated molecular or clinical networks underscores the fundamental interconnection between systemic factors and molecular mechanisms in atherosclerosis pathogenesis. By capturing both cohort-wide patterns through patient similarity networks and individual molecular perturbations through PPI networks, our approach affords an improved understanding of the multiscale nature of cardiovascular diseases where transcriptomics, clinical features, and molecular dysregulation converge to drive pathology. 

Importantly, ATHENA identifies molecularly informed clusters within clinically defined subtypes of atherosclerosis. Understanding these two clusters in each subtype addresses the complexity of disease pathology, indicating that patients grouped under a single clinical diagnosis may differ substantially at the molecular level. Establishing how these molecular phenotypes can meaningfully guide patient-specific treatment decisions and elucidate underlying disease mechanisms will necessitate careful clinical study design, targeted investigations, and subsequent validation in larger cohorts. Nevertheless, this novel stratification represents a critical step toward precision medicine approaches in atherosclerosis, potentially enabling earlier intervention and improved patient outcomes. 



\begin{acks}
We would like to acknowledge Baradwaj Simha Sankar for his feedback in guiding the direction of hierarchical analysis of atherosclerosis. This work was supported by National Institutes of Health (NIH) R35 HL135772 to P.P.; U34 HG012517 to P.P. and A.B.; and the TC Laubisch Endowment to P.P. at UCLA. 
\end{acks}

\bibliographystyle{ACM-Reference-Format}
\bibliography{sample-base}

\appendix

\end{document}